\documentclass[11pt]{article}
\usepackage{subfiles}
\usepackage{acl}

\usepackage{times}
\usepackage{latexsym}

\usepackage[utf8]{inputenc}

\usepackage{microtype}

\usepackage{todonotes}

\usepackage{xspace}
\usepackage{graphicx}
\usepackage{latexsym,amsmath}
\usepackage{amssymb}
\usepackage{blindtext}

\usepackage{times}
\newcommand{\quotes}[1]{``#1''}
\newcommand{\wdata}{EW10K}
\newcommand{\wdatah}{HW10K}

\newcommand{\wsms}{\mbox{\textsc{Warm-S}}\xspace}
\newcommand{\wsm}{\mbox{\textsc{Warm}}\xspace}
\newcommand{\lbf}{\mbox{\textsc{LBF}}\xspace}

\newcommand{\wsmwb}{\wsm\ w/o Beam Exploration}
\newcommand{\wsmwbs}{\wsms\ w/o Beam Exploration}

\title{WARM: A Weakly (+Semi) Supervised Math Word Problem Solver}

\author{ Oishik Chatterjee \thanks{\ \ \ The author contributed to this work while at IIT Bombay} \\ Flipkart \\ Bangalore India \And
        Isha Pandey \\ Department of CSE \\ IIT Bombay \\ \texttt{\{oishik75, iishapandey77, aashishwaikar\}@gmail.com} \And
        Aashish Waikar\footnotemark[1] \\ Quadeye \\ Gurgaon India 
        \AND
         Vishwajeet Kumar \\ IBM India Research Lab\\ Bangalore India \\ \texttt{ vishk024@in.ibm.com} \And
        Ganesh Ramakrishnan \\ Department of CSE \\ IIT Bombay\\\texttt{ganesh@cse.iitb.ac.in}}

\begin{document}
\maketitle
\begin{abstract}
 Solving math word problems (MWPs) is an important and challenging problem in natural language processing. Existing approaches to solve MWPs require full supervision in the form of intermediate equations. However, labeling every MWP with  its corresponding equations is a time-consuming and expensive task. In order to address this challenge of equation annotation, we propose a weakly supervised model for solving MWPs by requiring only the final answer as supervision. 
 We approach this problem by first learning to generate the equation using the problem description and the final answer, which we subsequently use to train a supervised MWP solver. We propose and compare various weakly supervised techniques to learn to generate equations directly from the problem description and answer. Through extensive experiments, we demonstrate that without using equations for supervision, our approach achieves accuracy gains of 4.5\% and 32\%
 over the state-of-the-art weakly supervised approach~\citep{hong2021learning}, on the standard Math23K~\citep{wang-2017-dns} and AllArith~\citep{roy-2017-unitdep} datasets respectively. 
 Additionally, we curate and release  new datasets of roughly 10k MWPs each in English and in Hindi (a low resource language).
These datasets are suitable for training weakly supervised models. We also present an extension of \wsm\footnote{\wsm\ stands for {\bf\underline{W}}e{\bf \underline{A}}kly supe{\bf\underline{R}}vised {\bf \underline{M}}ath solver.} to semi-supervised learning and present further improvements on results, along with insights. 
\end{abstract}

\section{Introduction}

A Math Word Problem (MWP) 
is a numerical problem expressed in natural language (problem description), that can be transformed into an equation (solution expression), which can be solved to obtain the final answer. In  Table 1, we present an example MWP. Automatically solving MWPs has recently gained lot of research interest in natural language processing (NLP). 
The task of automatically solving MWPs is challenging owing to two primary reasons: i) {\it The unavailability of large training datasets with problem descriptions, equations as well as corresponding answers} -- as depicted in Table 1,  equations can provide {\color{blue} full supervision}, since equations can be solved to obtain the answer, and the answer itself  amounts to {\color{brown} weak supervision} only; ii) {\it Challenges in parsing the problem description and representing it suitably for effective decoding of the equations}. Paucity of completely supervised training data can pose a severe challenge in training MWP solvers. Most existing approaches assume the availability of full supervision in the form of both intermediate equations and answers for training. However, annotating MWPs with equations is an expensive and time consuming task. There exists only two sufficiently large datasets ~\citep{wang-2017-dns} in Chinese and \citep{amini2019mathqa} in English consisting of MWPs with annotated intermediate equations for supervised training.







We propose a novel two-step weakly supervised technique to solve MWPs by making use only of the weak supervision, in the form of answers. In the first step, using only the answer as supervision, we learn to generate equations for questions in the training set. In the second step, we use the generated equations along with answers to train any state-of-the-art supervised model. We illustrate the effectiveness of our weakly supervised approach on our newly curated reasonably large dataset in English and a similarly curated dataset in Hindi - a low resource language. We also perform experiments with semi-supervision and demonstrate how our model can benefit from a small amount of completely labelled data.
Our main contributions are as follows: \\
\noindent 1) An approach, \wsm, ({\it c.f.},  Section~\ref{sec:our_approach}) for generating equations from MWPs, given (weak) supervision only in the form of the final answer. 

\noindent 2) An extended semi-supervised training method to leverage a small amount of annotated equations as strong/complete supervision.

\noindent 3) A new and relatively large dataset, \wdata, in English (with more than 10k instances), for training weakly supervised models for solving MWPs  ({\it c.f.},  Section~\ref{sec:dataset}). Given that weak supervision makes it possible to train MWP solvers even in the absence of extensive equation labels, we also present results on a similarly crawled dataset, \wdatah (with around 10k instances), in a low resource language, {\it viz.} Hindi, where we can avoid the additional effort required to generate equation annotations.  

\noindent 4) We empirically show that \wsm{} outperforms state-of-the-art models on  most of the datasets. Further, we empirically demonstrate the benefits of the semi-supervised extension to \wsm.

\includegraphics[scale=0.6]{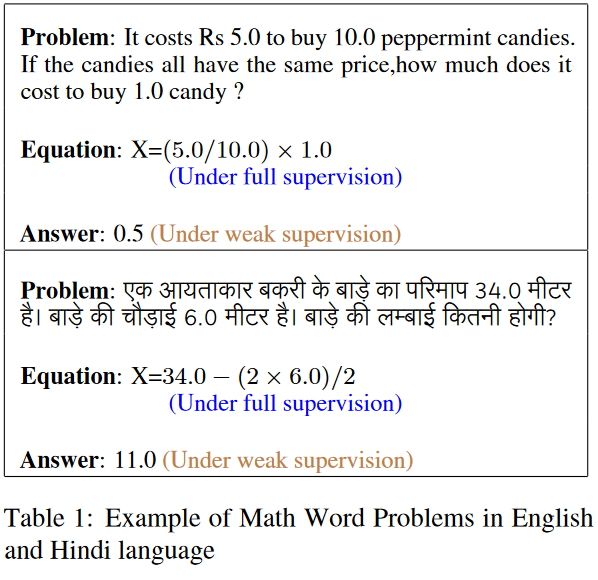}
\setcounter{table}{1}

\section{Related Work}
Automatic math word problem solving has recently drawn
significant interests in the natural language processing (NLP) community. 
Existing MWP solving methods can be broadly classified into four categories: (a) rule-based methods, (b) statistics-based methods, (c) tree-based methods, and (d) neural-network-based methods.

Rule-based systems~\citep{fletcher-1985-wordpro,bakman-2007-robust,yuhui-2010-frame} were amongst the earliest approaches to solve MWPs. They  rely heavily on hand-engineered rules that might cover a limited domain of problems. Statistics-based methods~\citep{hosseini-2014-learning,kushman-2014-learning,sundaram-2015-natural,mitra-2016-learning,liang-2016a-tag,liang-2016b-tag} use predefined logic templates and employ traditional machine learning models to identify entities, quantities, and operators from the problem text and subsequently employ simple logical inference to yield the numeric answer. \citep{upadhyay2016learning} employ a semi-supervised approach by learning to predict templates and corresponding alignments using both explicit and implicit supervision. Tree-based methods~\citep{roy-2015-genmwp,kedziorski-2015-parsing,roy-2016-eqpar,roy-2017-unitdep,roy-2018-mapmwp} replaced the process of deriving an equation by constructing an equivalent tree structure, step by step, in a bottom-up manner.

More recently, neural network-based MWP solving methods have been proposed~\citep{wang-2017-dns,wang-2018a-seq2seqet,wang-2018b-mathdqn,huang-2018-neural,chiang-2019-stack,wang-2019-trnn,liu-2019-tree,xie-2019-goal, wu2021math, shen2021generate}. These employ an encoder-decoder architecture and train in an end-to-end manner without the need for hand-crafted rules or templates. \citep{wang-2017-dns} were  the first to propose a sequence-to-sequence (Seq2Seq) model, {\it viz.}, Deep Neural Solver, for solving MWPs. They employ an RNN-based encoder-decoder architecture to directly translate the problem text into equation templates and also release a high-quality large-scale dataset, Math23K, consisting of 23,161 MWPs in Chinese.

\citep{liu-2019-tree} and \citep{xie-2019-goal} propose tree-structured decoding that generates the syntax tree of the equation in a top-down manner. In addition to applying tree-structured decoding, \citep{zhang-etal-2020-graph-tree} propose a graph-based encoder to capture relationships and order information among the quantities.
For a more comprehensive review on automatic MWP solvers, readers can refer to a recent survey paper~\citep{zhang-2018-survey}. 

Unlike all the previous works that require equations for supervision, \citep{hong2021learning} propose a weakly supervised method for solving MWPs, where the answer alone is required for training. Their approach attempts to  generate  the equation tree  in a rule based manner so that the correct answer is reached. They then train their model using the fixed trees.
With the same motivation. we also propose a novel weakly supervised model, \wsm, ({\it c.f.},  Section~\ref{sec:our_approach}) for solving MWPs using only the final answer for supervision. We show how \wsm{} can be extended to semi-supervised joint learning in the presence of weak answer-level supervision in conjunction with some equation-level supervision. Further, we  empirically demonstrate that \wsm{} outperforms~\citep{hong2021learning} on all the datasets.

We also took insights from ~\citep{kumar2018putting}, \citep{thakoor2018synthesis}, \citep{akula2021cross}, \citep{kumar2015machine}, \citep{singh2016building}, \citep{kumar2019cross}  and \citep{tarunesh2021meta} for handling mathematical data in two different languages.

This paper is organized as follows. In Section~\ref{sec:dataset}, we set the premise for our approach by describing the new datasets (\wdata{} and \wdatah) for weak supervision that we release. In Section~\ref{sec:our_approach}, we describe our weakly supervised approach \wsm{} and its semi-supervised extension \wsms{}. In Section~\ref{sec:expt}, we present the experimental setup whereas in Section~\ref{sec:results} we delve into the results and its analysis before concluding in Section~\ref{sec:conc}.

\section{Dataset}
\label{sec:dataset}
Currently, there does not exist any sufficiently large English dataset for single and simple 
equation MWPs.
While there exists an English dataset~\citep{amini2019mathqa} with sufficiently large MWPs, the questions in the dataset are meant to be evaluated in a multiple choice question (MCQ) manner. Also, the equation associated with each word problem in this dataset is significantly more complex and requires several binary and unary operators. On the other hand, Math23K~\citep{wang-2017-dns} is in Chinese and Dolphin18k~\citep{huang-etal-2016-well} contains mostly multi-variable word problems. To address these gaps, we curate a new English MWP dataset, {\it viz.},  \wdata{}\footnote[2]{\url{https://github.com/iishapandey/WARM}} consisting of 10227 word problem instances (each associated with a single equation) that can be used for training MWP solver models in a weakly supervised manner.

\begin{figure}[!ht]

\centering
\includegraphics[width=0.45\textwidth]{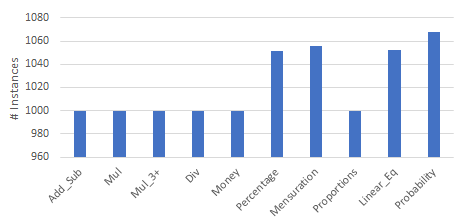}
\caption{Distribution on different types of questions}
\label{fig:type}
\end{figure}

We crawled IXL\footnote[3]{\url{https://in.ixl.com/}} to obtain MWPs for grades VI until X. These word problems involve a wide variety of mathematical computations ranging from simple addition-subtraction to much harder mensuration and probability problems. The dataset consists of 10 different types of problems, spanning 3 tiers of difficulty. We also annotate the dataset with the target unit. The exact distributions are presented in Figure~\ref{fig:type}.

We similarly created a MWP dataset in Hindi\footnotemark[2] - a low resource language. It consists of 9,896 question answer pairs. To the best of our knowledge, this is the first MWP dataset of such size in Hindi.

\section{Our Approach: \wsm}
\label{sec:our_approach}
We propose a weakly supervised model, \wsm{}, for solving the MWP using only the answer for supervision. It is a two-step cascaded approach for weakly supervised MWP solving. 
For the first step, we propose a model that predicts the equation, given a problem text and answer. This model uses reinforcement learning to search the space of possible equations,  given the question and the correct answer only. The answer acts as the goal of the agent and the search is terminated either when the answer is reached or when the equation length exceeds a pre-defined length (this is required, else the search space would be infinitely large). The model is designed to be a two layer bidirectional GRU~\citep{cho-etal-2014-learning} encoder and a decoder network with fully connected units (described in Section~\ref{decoder}). 
We refer to  this model as \wsm. Note that this model requires an answer to determine when to stop exploring. Since we ultimately want a model which should only take the problem statement as input and generate the answer (by generating the correct equation), this model alone is insufficient for evaluation. 
Using this model, we create a noisy equation-annotated dataset from the weakly annotated training dataset (the training dataset has answers since it is weakly supervised). We use only those instances to create the dataset for which the equation generated by the model yields the correct answer. Note that the equations are noisy, since there is no guarantee that the generated equation will be the shortest or even correct. In the second step, we use this noisy data for supervised training of a state-of-the-art model. The trained supervised model is finally used for evaluation. For simplicity, we provide a summary of notations in Section 1 in supplementary.

\begin{figure}[!ht]
\centering
\includegraphics[width=0.45\textwidth]{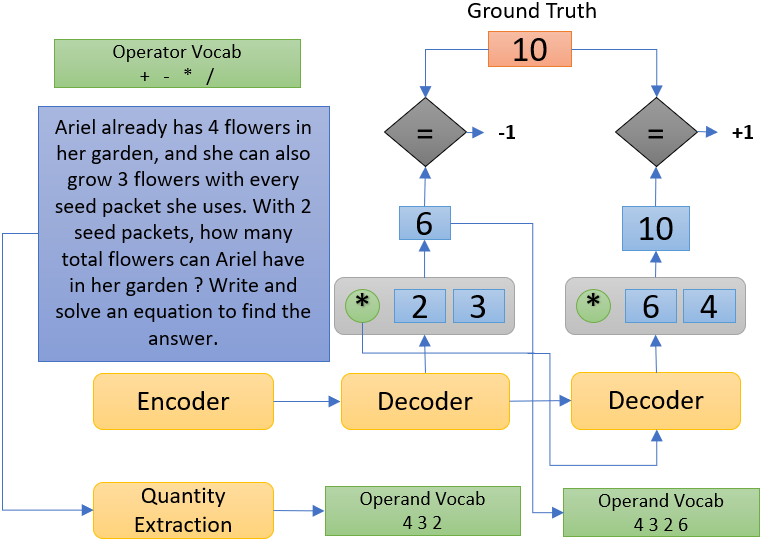}
\caption{Inference Illustration} 
\label{fig:inference}
\end{figure}

\begin{figure}[!htb]
\centering
\includegraphics[width=0.5\textwidth]{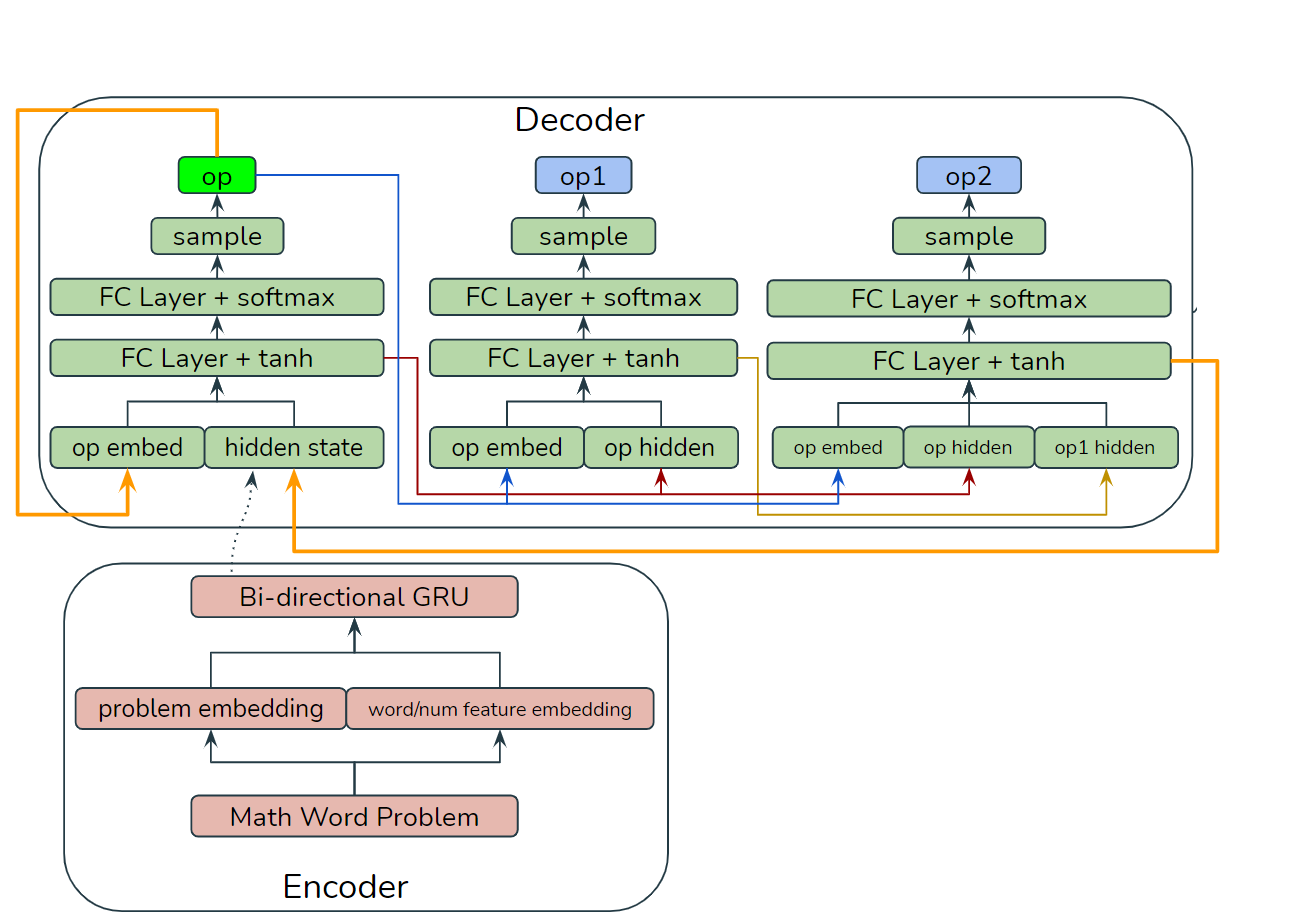}
\caption{Architecture for generating equation tree in \wsm.}
\label{fig:architecture}
\end{figure}

\subsection{Equation Generation} \label{eq:eqgen}
The first step of our approach is to generate equation given a problem text $P$ and answer $A$. This is done by using our \wsm model. The problem text is passed through the encoder of the \wsm model to get its encoded representation which is then fed to the decoder. At each time step, the decoder generates an operator and its two operands from the operator and operand vocabulary list. The operation is then executed to obtain a new quantity. This quantity is checked against the ground truth and if it matches the ground truth, the decoding is terminated and a reward of +1 is assigned. Else we assign a reward of -1 and the generated quantity is added to the operand vocabulary list and the decoding continues. The working of the \wsm model and architecture are illustrated in Figure~\ref{fig:inference} and Figure~\ref{fig:architecture} respectively. In the following few subsections, we describe the architecture as well as the training in details.

\subsection{Encoder} \label{sec:encoder}
The encoder takes as input, the MWP represented as a sequence of tokens $P = x_{1}x_{2}x_{3}...x_{n}$. We replace each number in the question with a special token $<num\_j>$ to obtain this sequence where $j$ denotes the index of number in the operand vocab for that question. Each word token $x_{i}$ is first transformed into the corresponding word embedding $\pmb{x}_i$ by looking up an embedding matrix $\pmb{M}_w$. Next, a binary feature is appended to the embedding to indicate whether the token is a word or a number. As depicted in the lower half of Figure~\ref{fig:architecture}, this appended embedding vector is then passed through a 2 layer bidirectional GRU~\citep{cho-etal-2014-learning} and the outputs from both directions of the final layer are summed to get the encoded representation of the text. This representation is then passed on to the decoder.

\subsection{Decoder}
\label{decoder}
The decoder consists of 3 fully connected networks for generating operator, left operand and the right operand. As illustrated in the upper half of Figure~\ref{fig:architecture}, the decoder takes as input the previous decoded operand and the last decoder hidden state and outputs the operator, left operand, right operand and hidden state at the current time step. We initialize the decoder hidden state with the last state of the encoder:

\begin{center} $o^{p}_{t}, o^{l}_{t}, o^{r}_{t}, h^{d}_{t} = DecoderFCN (o^{p}_{t-1}, h^{d}_{t-1})$ \end{center}

Here, $h^{d}_{t}$ is the decoder hidden state at the $t^{th}$ time step. $o^{p}_{t}$, $o^{l}_{t}$ and $o^{r}_{t}$ are probability distributions over operators, left and right operands respectively.

\subsubsection{Operator generation}
Inside our decoder, we learn an operator embedding matrix $Em_{op}(op_{t-1})$, where $op_{t-1}$ is the operator sampled in the last time step. We generate the operator hidden state $h^{op}_{t}$ using a gating mechanism.
\begin{equation*}
    g^{op}_{t} = \sigma(W^{1}_{op} [Em_{op}(op_{t-1}) ; h^{d}_{t-1}] + b^{1}_{op})
\end{equation*}
\begin{equation*}
    h^{op}_{t} = g^{op}_{t} * tanh(W^{2}_{op} [Em_{op}(op_{t-1}) ; h^{d}_{t-1}] + b^{2}_{op})
\end{equation*}
\begin{equation*}
    o^{p}_{t} = softmax(W^{3}_{op} h^{op}_{t} + b^{3}_{op}) 
\end{equation*}
Here $\sigma()$ denotes the sigmoid function and $*$ denotes elementwise multiplication. We sample operator $op_{t}$ from the probability distribution $o^{p}_{t}$.

\subsubsection{Left Operand Generation}
We use the embedding of the current operator $Em(op_{t})$ and the operator hidden state $h^{op}_{t}$ to obtain a probability distribution over the operands. We employ a similar gating mechanism as used for generating \textit{operator}.
\begin{equation*}
    g^{ol}_{t} = \sigma(W^{1}_{ol} [Em_{op}(op_{t}) ; h^{op}_{t}] + b^{1}_{ol})
\end{equation*}
\begin{equation*}
    h^{ol}_{t} = g^{ol}_{t} * tanh(W^{2}_{ol} [Em_{op}(op_{t}) ; h^{op}_{t}] + b^{2}_{ol})
\end{equation*}
\begin{equation*}
    o^{l}_{t} = softmax(W^{3}_{ol} h^{ol}_{t} + b^{3}_{ol}) 
\end{equation*}
We sample the left operand $ol_{t}$ from the probability distribution $o^{l}_{t}$.

\subsubsection{Right Operand Generation}
For generating the right operand, we use the additional context information that is already available from the generated left operand. Thus, in addition to the operator embedding $Em_{op}(op_{t})$ and operator hidden state $h^{op}_{t}$ we also use the left operand hidden state to get the right operand hidden state $h^{or}_{t}$.
\begin{equation*}
    g^{or}_{t} = \sigma(W^{1}_{or} [Em_{op}(op_{t}) ; h^{op}_{t} ; h^{ol}_{t}] + b^{1}_{or})
\end{equation*}
\begin{equation*}
    h^{or}_{t} = g^{or}_{t} * tanh(W^{2}_{or} [Em_{op}(op_{t}) ; h^{op}_{t} ; h^{ol}_{t}] + b^{2}_{or})
\end{equation*}
\begin{equation*}
    o^{r}_{t} = softmax(W^{3}_{or} h^{or}_{t} + b^{3}_{or}) 
\end{equation*}
We sample the right operand $or_{t}$ from the probability distribution $o^{l}_{t}$. The hidden state $h^{or}_{t}$ is returned as the current decoder state $h^{d}_{t}$.

\subsubsection{Bottom-up Equation Construction}
For each training instance, we maintain a dictionary of possible operands $OpDict$. Initially, this dictionary contains the numeric values from the instance, {\it i.e.}, the number tokens we have replaced with $<num\_j>$ during encoding. At the $t^{th}$ decoding step, we sample an operator $op_{t}$, left operand $ol_{t}$ and right operand $or_{t}$. We get an intermediate result by using the operator corresponding to $op_{t}$ on the operands $ol_{t}$ and $or_{t}$. This intermediate result is added to $OpDict$ which enables us to reuse the results of previous computations in future decoding steps. Thus, $OpDict$ acts as a dynamic dictionary of operands and we use it to progress towards the final answer in a bottom-up manner.

\subsection{Rewards and Loss} \label{eq:rewards}

We use the REINFORCE~\citep{10.1007/BF00992696} algorithm for training the model using just the final answer as the ground truth. We model the reward as $+1$ if the predicted answer matches the ground truth and $-1$ if the predicted answer does not equal the ground truth.
\par
Let $R_t$ be defined as the reward obtained after generating $y_t = (op_t,o_l,o_r)$. The probability $P_t$ of generating the tuple $y_t$ is specified by $p_\theta(y_t) = \prod \limits _{i=1}^t o^p_i \times o^l_i \times o^r_i$. 
The loss is specified  as  $L = - \sum \limits _i \mathbb{E}_{p_\theta(y_i)} [R_i]$ and the corresponding gradient is $\nabla_\theta L = \sum \limits _i \sum \limits _{y_i} p_\theta(y_i) R_i \nabla_\theta \log p_\theta(y_i) $.

Since the space of $y_i$ makes it infeasible to compute the exact gradient, we use the standardized technique of sampling $y_i$ from $p_\theta(y_i)$ to obtain an estimate  of the gradient.

\subsection{Beam Exploration in Training}
\label{sec:beam_search}

Since the reward space for our problem is very sparse, we observe that during model training, the gradients  go to zero. Our model converges too quickly to some local optima and consequently, the training accuracy saturates to some fixed value despite performing training for a large number of epochs. In order to counter this problem, we employ beam exploration in the training procedure. Instead of sampling operator $op_{t}$, left operand $ol_{t}$ and right operand $or_{t}$ only once in each decoding step, we sample $w$ triplets $(op_{t},ol_{t},or_{t})$ without replacement from the joint probability space in each decoding step. Here $w$ is the beam width. This helps in exploring $w$ different paths each epoch, thus increasing the exploration capabilities and reduce the problem of  cold start. In order to select beams from all possible candidates, we have tried multiple heuristics by inspecting the probability and reward values. We have observed empirically that selecting the beam that gives a positive reward at the earliest decoding step yields the best performance.  This enables our model to explore more and mitigates the above problem significantly. 

\subsection{\wsms: Adding Semi-supervision} \label{sec:warms}
While it is expensive to completely label large MWP datasets  with equations, it is relatively easier to annotate a small percentage of that data. We argue that addition of this small amount of semi-supervision can improve the model training significantly.
\par
We, therefore, consider a model that benefits from a relatively small amount of \textit{strong} supervision in the form of equation annotated data: $D_s$, in addition to a potentially larger sized math problem datasets with only \textit{weak} supervision $D_w$. For a data instance $d$: $d.p$, $d.e$, and $d.a$ represent its problem statement, equation, and answer respectively. $D_s$ consists of instances of the form $(d.p, d.e, d.a)$ while $D_w$ contains instances of the form $(d.p, d.a)$. We extend the \wsm\ model to include a Cross-Entropy loss component for instances belonging to $D_s$. The net loss is the sum of the REINFORCE ($RL_{\wsm}$) and Cross-Entropy losses shown below:-

\hspace{-12 pt}\textbf{Loss 1}:$\sum\limits_{d \in D_w}RL_{\wsm}(d.p, d.a)$ 

\hspace{-12 pt}\textbf{Loss 2}: $\sum \limits _{d \in D_s}Cross\_Entropy(d.e,$\\

\vspace{-15 pt}

\hspace{4.45cm}$\wsm(d.p, d.a))$

Thus, we facilitate semi-supervision through \textbf{Loss 2}. 
That is, we jointly use the equations predicted (by \wsm) for datapoints belonging to $D_w$ and the ground truth equations for instances belonging to $D_s$, for  training any state-of-the-art supervised MWP solver.

\section{Experimental Setup} \label{sec:expt}

In this section, we report details of the experiments  on four datasets to examine the performance of the proposed weakly supervised model \wsm{} and its semi-supervised extension \wsms{}. We present comparisons with various baselines as well as with fully supervised models.
\subsection{Datasets}
We perform all our experiments on the publicly available AllArith~\citep{roy-2017-unitdep} and Math23K~\citep{wang-2017-dns} datasets and also on our \wdata{} and \wdatah{} datasets.For each dataset, we have used a $80:20$ train-test split.\\
\noindent {\textbf{AllArith}} contains 831 MWPs, annotated with equations and answers. It is populated by collecting problems from smaller datasets, {\it viz.}, \textbf{AI2}~\citep{hosseini-2014-learning}, \textbf{IL}~\citep{roy-2015-genmwp}, \textbf{CC}~\citep{roy-2015-genmwp} and \textbf{SingleEQ}~\citep{kedziorski-2015-parsing}. All mentions of quantities are  normalized to digits. Further, near-duplicate problems (with over 80\% match of unigrams and bigrams) are filtered out. \\
\noindent {\textbf{Math23K}}~\citep{wang-2017-dns} contains 23,161 MWPs in Chinese with 2187 templates, annotated with equations and answers, for elementary school students and is crawled from multiple online education websites. It is the largest publicly available dataset for the task of automatic MWP solving.\\
\noindent {\textbf{\wdata}} ({\it c.f.}, Section \ref{sec:dataset}) contains 10,227 MWPs in English and {\textbf{\wdatah}} contains 9,896  in Hindi for classes VI to X. We employ a $80:20$ train-test split in each case.

\subsection{Dataset Preprocessing}
\label{sec:preprocess}
We replace every number token in the problem text with a special word token $<num\_j>$ before providing it as input to the encoder. We also define a set of numerical constants $V_{const}$ to solve those problems which might require special numeric values that may not be present in the problem text. For example, consider the problem \quotes{The radius of a circle is 2.5, what is its area?}, the solution is \quotes{$\pi$ x 2.5 x 2.5}, but the constant quantity $\pi$ cannot be found in the text. As our model does not use equations as supervision, we cannot know precisely what extra numeric values might be required for a problem, so we fix $V_{const} = \{1, \pi \}$. Finally, the operand dictionary for every problem is initialised as $OpDict = n_P \cup V_{const}$ where $n_P$ is the set of numeric values present in the problem text.

\subsection{Implementation Details}

We implement\footnote[4]{Source code is attached as supplementary material} 
all our models in PyTorch~\citep{paszke2019pytorch}. We set the dimension of the word embedding layer to 128, and the dimension of the hidden states for other layers to 512. We use the REINFORCE~\citep{10.1007/BF00992696} algorithm and Adam~\citep{article} to optimize the parameters. The initial value of the learning rate is set to 0.001, and the learning rate is multiplied by 0.7 every 75 epochs. We also set the dropout probability to 0.5 and weight decay to 1e-5 to avoid over-fitting. Finally, we set the beam width to 5 in beam exploration.
We train our model for 200 epochs with the batch size set to 256.

\subsection{Models}
We compare the MWP solving accuracy of our weakly supervised models with beam exploration
on  the following set of baseline and fully supervised models: 

\noindent \textbf{\wsm}  
is the proposed weakly supervised approach to equation generation (described from Section~\ref{eq:eqgen} until~\ref{eq:rewards}) 
by employing beam exploration ({\it c.f.}, Section \ref{sec:beam_search}).\\ 
\noindent \textbf{\wsmwb} is  \wsm\ without beam exploration while decoding.\\
\noindent \textbf{\wsms} is the semi-supervised extension to \wsm{} ({\it c.f.}, Section~\ref{sec:warms}) 
using beam exploration (Section \ref{sec:beam_search}).

\noindent \textbf{\wsmwbs} is the same as \wsms{} but does not use beam exploration while decoding.\\
\noindent \textbf{Random Equation Sampling} consists of a random search over $k$ parallel paths of length $d$. For each path, an operator and its two operands are uniformly sampled from the given vocabulary and the result is added to the operand vocabulary (similar to \wsm). The equation is terminated once the correct answer is reached. We set $k=5$ and $d=40$ for a fair comparison with our model in terms of the number of search operations.\\
\noindent \textbf{Seq2Seq Baseline} is a GRU~\citep{cho-etal-2014-learning} based seq2seq encoder-decoder model. REINFORCE~\citep{10.1007/BF00992696} is used to train the model. Beam exploration is also employed to mitigates issues mentioned in Section~\ref{sec:beam_search}.\\
\noindent \textbf{LBF} \citep{hong2021learning} is a weakly supervised model which uses only answer as supervision by fixing incorrect equation parse trees in each iteration. It subsequently performs training with the fixed trees.\\
\noindent \textbf{Hybrid model w/ SNI}~\citep{wang-2017-dns} is a combination of the retrieval  and the RNN-based Seq2Seq models with significant number identification (SNI). \\
\noindent \textbf{Ensemble model w/ EN}~\citep{wang-2018a-seq2seqet} is an ensemble model that selects the result according to generation probability across Bi-LSTM, ConvS2S and Transformer Seq2Seq models with equation normalization (EN).\\
\noindent \textbf{Semantically-Aligned}~\citep{chiang-2019-stack} is a Seq2Seq model with an encoder designed to understand the semantics of the problem text and a decoder equipped with a stack to facilitate tracking the semantic meanings of the operands.\\
\noindent \textbf{T-RNN + Retrieval}~\citep{wang-2019-trnn} is a combination of the retrieval model and the T-RNN model comprising a structure prediction module that predicts the template with unknown operators and an answer generation module that predicts the operators.\\
\noindent \textbf{Seq2Tree}~\citep{liu-2019-tree} is a Seq2Tree model with a Bi-LSTM encoder and a top-down hierarchical tree-structured decoder consisting of an LSTM that makes use of the parent and sibling information fed as the input.

\noindent \textbf{GTS}~\citep{xie-2019-goal} is a tree-structured neural model that generates the expression tree in a goal-driven manner.

\noindent \textbf{Graph2Tree}~\citep{zhang-etal-2020-graph-tree} consists of a graph-based encoder which captures the relationships and order information among the quantities. It also employs a tree-based decoder that generates the expression tree in a goal-driven manner. 

As described earlier in Section~\ref{sec:our_approach}, we use our weakly supervised models (\wsm{} and \wsms) to generate labelled data ({\it i.e.}, equations) which we then use to train a supervised model. We have performed experiments using GTS~\citep{xie-2019-goal} and Graph2Tree~\citep{zhang-etal-2020-graph-tree} as the supervised models since they are the current state-of-the-art.

\section{Results and Analysis}
\label{sec:results}

\tabcolsep=0.11cm
\renewcommand{\arraystretch}{1.2}

\begin{table}[!htb]
    \centering
    \small
    \scalebox{0.8}{
        \begin{tabular}{| l | c | c | c | c |}
        \hline
        \textbf{Weakly Supervised Models} & AllArith & Math23K & \wdata & \wdatah\\ \hline
            \wsmwb & 42.1 & 14.5 &  57.5 & 67.3\\
            \wsm & \textbf{97.4} & \textbf{93.8} & \textbf{99.3} & \textbf{99.5}\\ 
            \hline
        \textbf{Baselines} & AllArith & Math23K &  \wdata & \wdatah\\ \hline
            Random Equation Sampling &  53.4 &  17.6 & 46.3 & 66.6\\
            Seq2Seq Baseline &  67.0 & 7.1  & 77.6 & 75.8\\
            \hline
        \end{tabular}
    }
    \caption{Equation generation accuracies of \wsm{} based models compared to baselines. All models are trained using ground truth answers on the training set. \wsm outperforms all the remaining models by as significant margin on all the datasets. Evidently, beam exploration significantly improves performance.}
    \label{tab:res_eqngen}
\end{table}

\begin{table}[!htb]
    \centering
    \small
    \scalebox{0.69}{
        \begin{tabular}{| l | c | c | c | c |c|}
        \hline
        \textbf{Weakly Supervised Models} & AllArith & Math23K & \wdata & \wdatah\\ \hline
            \wsmwb (GTS) & 36.1 & 12.8 & 52.6 & 54.1\\
            \wsm (GTS) & 66.9 & 55.3 &  86.9 & 81.5\\ 
            \wsmwb (Graph2Tree) & 48.2 & 13.5 & 49.8 & 58.3\\
            \wsm (Graph2Tree) & \textbf{68.7} & \textbf{56.0} &  \textbf{87.2} & \textbf{82.9}\\ 
            \lbf$^\ddagger$ & 51.8 & 53.6 & 81.3 & 75.8\\


            \hline
        \textbf{Fully Supervised Models} & AllArith & Math23K & \wdata & \wdatah\\ \hline
            Graph2Tree$^\ddagger$ & 71.9 & 75.5 & NA & NA\\
            GTS$^\ddagger$ & 70.5 & 73.6 & NA & NA\\
            Seq2Tree & -- & 69.0 & NA & NA\\
            T-RNN + Retrieval & -- & 68.7 & NA & NA\\ 
            Semantically-Aligned$^\dagger$ & -- & 65.8 & NA & NA\\
            Ensemble model w/ EN & -- & 68.4 & NA & NA\\
            Hybrid model w/ SNI$^\dagger$ & -- & 64.7 & NA & NA\\
            \hline
        \end{tabular}
    }
    \caption{MWP solving accuracy of \wsm{}-based models compared to various supervised models on AllArith and Math23K datasets. $\dagger$ denotes that result was reported on  5-fold cross validation. All other models are tested on the test set. $\ddagger$ denotes that the result is on the same train-test split as ours. \quotes{--} denotes code unavailability/reproducibility issues. NA is not applicable. }
    \label{tab:res_acc}
\end{table}

\begin{table}[]
\small
\scalebox{0.85}{
\begin{tabular}{|p{0.5\textwidth}|}
\hline
\textbf{Problem}: Ariel already has 4.0 flowers in her garden, and she can also grow 3.0 flowers with every seed packet she uses. With 2.0 seed packets, how many total flowers can Ariel have in her garden ?                                 \\ \hline
\textbf{Answer}: 10.0                                                                                                                                                                                                                                                                  \\ \hline
\textbf{Equation Generated}: X=(4.0+(2.0*3.0))  (Correct)                                                    
\\ \hline \hline

\textbf{Problem}: Celine took a total of 6.0 quizzes over the course of 3.0 weeks. After attending 7.0 weeks of school this quarter, how many quizzes will Celine have taken in total ? Assume the relationship is directly proportional. \\ \hline
\textbf{Answer}: 14.0                                                                                                                                                                                                                                                                   \\ \hline
\textbf{Equation Generated}: X=(7.0+7.0) (Incorrect)                                                                \\ \hline

\end{tabular}}
\caption{Equation Generated by \wsm  model}
\label{tab:eq_gen}

\end{table}

\begin{table}[]
\small
\scalebox{0.85}{
\begin{tabular}{|p{0.5\textwidth}|}
\hline



\textbf{Problem}: Latrell ordered a set of yellow and purple pins. He received 72.0 yellow pins and 8.0 purple pins. What percentage of the pins were yellow?
\\ \hline
\textbf{Equation Generated by WARM (G2T)}: X=(72.0*(100.0/(72.0+8.0)))(Correct)                     
\\ \hline
\textbf{Equation Generated by LBF}: X=(1.0+(1.0+72.0)) (Incorrect)
\\ \hline \hline

\textbf{Problem}: A square barn has a perimeter of 28.0 metres. How long is each side of the barn ? \\ \hline
\textbf{Equation Generated by WARM(G2T)}: X=((28.0/2.0)/2.0) (Correct)                                                                                                                                                                                                  \\ \hline
\textbf{Equation Generated by LBF}: X=((28.0+28.0)/28.0) (Incorrect)                                                                        \\ \hline                                                        
\end{tabular}}
\caption{Comparing \wsm and LBF model predicted equations}
\label{tab:cmp_eq_gen}
\end{table}

\subsection{Analyzing \wsm}
In Table~\ref{tab:res_eqngen}, we observe that our model \wsm yields far higher accuracy than random baselines with the accuracy values close to 100\% on AllArith and \wdata. Thus we are able to more accurately generate equations for a given problem and answer which can then be used to train supervised models. Please note that, in Table~\ref{tab:res_eqngen}, we report equation generation accuracies on the training set by training the weakly supervised and baseline models using ground truth answers on the training set. 

As has been discussed earlier in Section \ref{sec:beam_search}, our model \wsmwb\ suffers from the problem of converging to local optima because of the sparsity of the reward signal. Training our weakly supervised models with beam exploration alleviates the issue to a large extent as we explore the solution space much more extensively and thus partly circumventing the sparsity issue. We observe vast improvement in the training accuracy by introduction of beam exploration. The model \wsm{} yields training accuracy significantly higher than its non-beam-explore counterpart. \wsm{} yields the best training accuracy overall. Since the equation generation accuracies of the baselines reported in Table~\ref{tab:res_eqngen} are far worse,the MWP solving accuracies turn out to be significantly worse - around 8-10\%, and hence we do not report them.

We also observe that \wsm{} yields results comparable to the various supervised models without requiring any supervision from gold equations. On AllArith, \wsm{} achieves an accuracy of 66.9\% and 68.7\% using GTS and Graph2Tree as the supervised models respectively. The state-of-the-art supervised model Graph2Tree yields 71.9\%. On Math23k, the difference between \wsm{} and the supervised models is more pronounced. \wsm's performance is comparable to that of \lbf on Math23k but significantly better on AllArith, \wdata and \wdatah, as evident in Table~\ref{tab:res_acc}. We have shown a comparison of predicted equations by \lbf and \wsm{} in Table~\ref{tab:cmp_eq_gen} 

In Table~\ref{tab:eq_gen}, we present some predictions. As can be seen, the model is capable of producing long complex equations as well. Sometimes, it may reach the correct answer but through an incorrect equation. {\it E.g.}: In the last example, the correct equation would have been $X=7.0*6.0/3.0$, but the model predicted $X=7.0+7.0$.

\subsection{Analysing Semi-supervision through \wsms}
For analyzing semi-supervision, we combined AllArith (831) with \wdata{} (10227). We randomly sampled 80\% of this data (8846) as our train-set. In retrospect, our train-set consists of 560 instances from AllArith that are completely labelled (amounting to 6.3\% of the train-set). We compare our semi-supervised approach against the weakly supervised approach, wherein the entire training data is treated as having only answer labels. 
\par
In Table~\ref{tab:res_eqngen_semisup}, we observe that with less than 10\% of fully annotated data, our equation exploration accuracy increases from 56.7\% to 92.0\% without beam exploration and 99.0\% to 99.2\% with beam exploration. In Table~\ref{tab:res_acc_semisup}, we also observe a similar trend while training the supervised models; our final MWP solving accuracy increases from 51.2\% to 87.4\% for \wsmwb{} and Graph2Tree as the supervised model. We also study thethere effect of varying amount of complete supervision in Supplementary Section:2.
\tabcolsep=0.11cm
\renewcommand{\arraystretch}{1.2}

\begin{table}[!htb]
    \centering
    \small
    \scalebox{0.98}{
        \begin{tabular}{| l | c | c |}
        \hline
        \textbf{Weakly Supervised Models} & AllArith +\wdata \\ \hline
        \wsmwb & 56.7\\
        \wsm & 99.0\\ \hline
        \textbf{Semi Supervised Models} & AllArith+\wdata\\ \hline
        \wsmwbs &  92.0\\
        \wsms &  99.2\\ \hline
        
        \end{tabular}
    }
    \caption{Equation generation accuracy of \wsms{} compared to weakly supervised models and baselines.}
    \label{tab:res_eqngen_semisup}
\end{table}
\begin{table}[!htb]
    \centering
    \small
    \scalebox{0.87}{
        \begin{tabular}{| l | c | c |}
        \hline
        \textbf{Weakly Supervised Models} & AllArith +\wdata\\ \hline
            \wsmwb (GTS) &  50.2\\
            \wsm (GTS) & 87.2\\
            \wsmwb (Graph2Tree) & 51.2\\
            \wsm (Graph2Tree) & 87.8\\
            \hline
         \textbf{Semi Supervised Models} & AllArith+\wdata\\ \hline
         \wsmwbs (GTS) & 87.2\\
         \wsms (GTS) & 92.1\\ 
         \wsmwbs (Graph2Tree) & 87.4\\
         \wsms (Graph2Tree) & 93.6\\ \hline
        \end{tabular}
    }
    \caption{MWP solving accuracy of \wsms{} compared to \wsm{}. With semi-supervision, there is a significant increase in accuracy for \wsmwb, bringing its performance closer to \wsm. }
    \label{tab:res_acc_semisup}
\end{table}

\section{Conclusion}
\label{sec:conc}
We have proposed a two step approach to solving math word problems, using only the final answer for supervision. Our weakly supervised approach, \wsm,  achieves a reasonable accuracy of 56.0 on the standard Math23K dataset even without leveraging equations for supervision. We also curate and release large scale MWP datasets, \wdata, in English and \wdatah, in Hindi. 
We observed that the results are encouraging for simpler MWPs. We also present the benefits of incorporating a semi-supervised extension to \wsm{}.

\section*{Acknowledgements}
We thank anonymous reviewers and  Rishabh Iyer for providing constructive feedback and suggestions.
Ganesh Ramakrishnan is grateful to IBM Research, India (specifically the IBM AI Horizon Networks - IIT Bombay initiative) as well as the IIT Bombay Institute Chair Professorship for their support and sponsorship.
We would like to acknowledge Saiteja Talluri and Raktim Chaki for their contributions in the initial stages of the work.

\bibliography{anthology,acl2021}
\bibliographystyle{acl_natbib}


\appendix

\section{Appendix}
\label{sec:appendix}


\subsection{Notations}
We summarize the notations used in section 4 of the main paper in table \ref{tab:not}.
\begin{table}[h]
\scalebox{0.65}{
\centering
\small

\begin{tabular}{|c|c|}
\hline
\textbf{Notation}          & \textbf{Description}                                                  \\ \hline

$W$ & Weight of the FC layers.
\\ \hline
$o^p_t$    & Probability distribution of operators at decoding timestep t.         \\ \hline
$o^l_t$    & Probability distribution of the left operand at decoding timestep t.  \\ \hline
$o^r_t$    & Probability distribution of the right operand at decoding timestep t. \\ \hline
$h^d_t$    & Decoder hidden state at timestep t.                                   \\ \hline
$Em_{op}$                     & Operator Embedding Matrix                                                      \\ \hline
$h^{op}_t$ & Hidden state for the operator at timestep t                           \\ \hline
$op_t$                      & Operator sampled from $o^p_t$.                         \\ \hline
$h^{ol}_t$   & Hidden state for the left operand at timestep t.                      \\ \hline
$ol_t$                      & Left operand sampled from $o^l_t$.                     \\ \hline
$h^{or}_t$   & Hidden state for the right operator at timestep t.                    \\ \hline
$or_t$                      & Right operator sampled from $o^r_t$.                   \\ \hline
OpDict                     & Operand dictionary used while decoding                                \\ \hline
$R_t$                       & Rewards obtained at timestep t.                                       \\ \hline
$p_\theta(y_t)$             & Probability of generating $y_t=(op_t, ol_t, or_t)$ at timestep t.   \\ \hline
\end{tabular}
}
\caption{Summary of notation used.}
\label{tab:not}
\end{table}

\subsection{Ablation Study: Varying Amount of Semi-supervision}
We performed an experiment to study the effect of different amounts of supervision by varying the number of instances in training set we treat as fully labelled. The number of fully labelled instances is X-axis*80. We observe that just having 160 equation-labelled instances (out of 8846 ie. 1.8\%) improves the equation-exploration accuracy significantly (46.7\% to 90.6\%) when we don't use beam exploration.

\begin{figure}[!htb]

\centering
\includegraphics[width=0.5\textwidth]{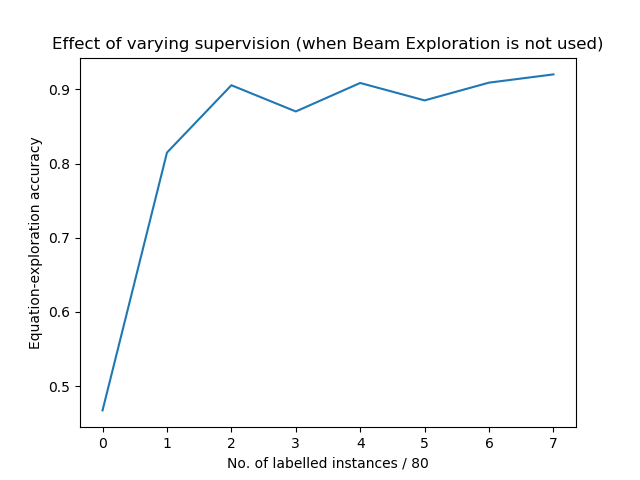}
\caption{Equation Exploration accuracy with varying supervision}
\label{fig:ablation}
\end{figure}

\subsection{Infrastructre Details}
\noindent\textbf{GPU Model used} :\\
\noindent 1)Model number: GeForce GTX 1080 Ti\\
\noindent 2)Memory : 12GB\\

\noindent\textbf{Training time} :\\
\noindent 1) WARM takes 4 hours for training\\
\noindent 2) G2T takes 1 hour and 30 minutes to get trained completely \\

\end{document}